# Clustering belief functions based on attracting and conflicting metalevel evidence

Johan SCHUBERT

Department of Data and Information Fusion
Division of Command and Control Systems
Swedish Defence Research Agency
SE-172 90  Stockholm, Sweden
`schubert@foi.se`
`http://www.foi.se/fusion/`

**Abstract**

In this paper we develop a method for clustering belief functions based on attracting and conflicting metalevel evidence. Such clustering is done when the belief functions concern multiple events, and all belief functions are mixed up. The clustering process is used as the means for separating the belief functions into subsets that should be handled independently. While the conflicting metalevel evidence is generated internally from pairwise conflicts of all belief functions, the attracting metalevel evidence is assumed given by some external source.

**Kewords:** belief functions, Dempster-Shafer theory, clustering.

## 1  Introduction

In this paper we extend an earlier method within Dempster-Shafer theory [8] for handling belief functions that concern multiple events. This is the case when it is not known a priori to which event each belief function is related. The belief functions are clustered into subsets that should be handled independently.

Previously, we developed methods for clustering belief functions based on their pairwise conflict [2, 6]. These conflicts were interpreted as metalevel evidence about the partition of the set of belief functions [4]. Each piece of conflicting metalevel evidence states that the two belief functions do not belong to the same subset.

The method previously developed is here extended into also being able to handle the case of attracting metalevel evidence. Such evidence is not generated internally in the same way as the conflicting metalevel evidence. Instead, we assume that it is given from some external source as additional information about the partitioning of the set of all belief functions.

For example, in intelligence analysis we may have conflicts (metalevel evidence) between two different intelligence reports about sighted objects, indicating that two objects probably does not belong to the same unit (subset). At the same time we may have information from communication intelligence as an external source (providing attracting metalevel evidence), indicating that the two objects probably do belong to the same unit (subset) as they are in communication.

We begin (Section 2) by giving an introductory problem description. In Section 3 we interpret the meaning of attracting and conflicting metalevel evidence. We assign values to all such pieces of evidence. In Section 4 we combine the metalevel evidence separately for each subset. Here, all attracting metalevel evidence, and all conflicting metalevel evidence are combined as two independent combinations within each subset. At the partition level (Section 5) we combine all metalevel evidence from the subsets, yielding basic beliefs for and against the adequacy of the partition. In Section 6 we compare the information content of attracting metalevel evidence with conflicting metalevel evidence. This is done in order to find a weighting between the basic beliefs for and against the adequacy of the partition in the formulation of a metaconflict function. The order of processing is shown in Figure 1. Finally, the metaconflict function is minimized as the method of finding the best partition of the set of belief functions (Section 7).

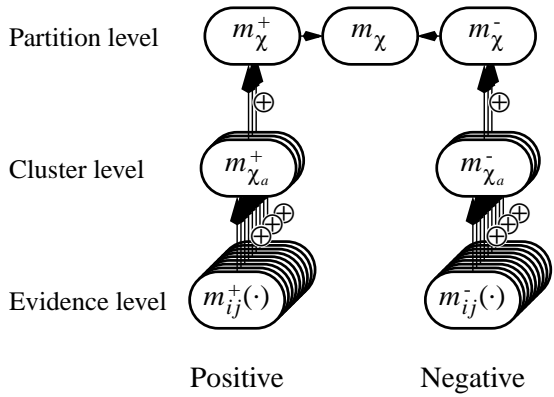

Figure 1: Order of processing.

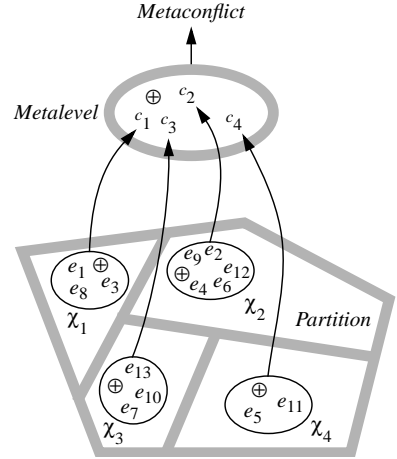

Figure 2: The conflict in each subset is interpreted as evidence at the metalevel.

## 2 Problem description

When we have several belief functions regarding different events that should be handled independently we want to arrange them according to which event they are referring to. We partition the set of belief function $\chi$ into subsets where each subset $\chi_i$ refers to a particular event, Figure 2. The conflict of Dempster's rule when all belief functions in $\chi_i$ are combined is denoted $c_i$. In Figure 2, thirteen belief functions $e_i$ are partitioned into four subsets. As these events have nothing to do with each other, they should be analyzed independently.

If it is uncertain whether two different belief functions are referring to the same event we do not know if we should put them into the same subset or not. We can then use the conflict of Dempster's rule when the two belief functions are combined, as an indication of whether belong together. A high conflict between the two functions is an indication of repellence that they do not belong together in the same subset. The higher this conflict is, the less credible that they belong to the same subset. A zero conflict, on the other hand, is no indication at all.

For each subset we may create a new belief function on the metalevel with a proposition that we do not have an "adequate partition." The new belief functions does not reason about any of the original problems corresponding to the subsets. Rather they reason about the partition of the other belief functions into the different subsets. Just so we do not confuse the two types of belief functions, let us call the new ones "metalevel evidence" and let us say that their combination take place at the metalevel, Figure 2.

On the metalevel we have a simple frame of discernment where $\Theta = \{\text{AdP}, \neg\text{AdP}\}$, where AdP is short for "adequate partition." Let the proposition take a value equal to the conflict of the combination within the subset,

$$m_{\chi_i}(\neg\text{AdP}) \overset{\Delta}{=} \text{Conf}(\{e_j | e_j \in \chi_i\})$$

where $\text{Conf}(\{e_j | e_j \in \chi_i\})$ is the conflict of Dempster's rule when combining all basic probability assignments in $\chi_i$.

In [4] we established a criterion function of overall conflict for the entire partition called the metaconflict function (Mcf). The metaconflict is derived as the plausibility of having an adequate partitioning based on $\oplus \{m_{\chi_i}(\neg AdP)\}$ for all subsets $\chi_i$.

DEFINITION. *Let the* metaconflict function,

$$Mcf(\{e_1, e_2, ..., e_n\}) \stackrel{\Delta}{=} 1 - \prod_{i=1}^{r}(1 - c_i),$$

*be the conflict against a partitioning of n belief functions of the set $\chi$ into r disjoint subsets $\chi_i$.*

Minimizing the metaconflict function was the method of partitioning the belief functions into subsets representing the different events.

However, instead of considering the conflict in each subset we may refine our analysis and consider all pairwise conflicts between the belief functions in $\chi_i$ [6], $m_{ij}^-(\cdot) = c_{ij}$, where $c_{ij}$ is the conflict of Dempster's rule when combining $e_i$ and $e_j$. When $c_{ij} = 1$, $e_i$ and $e_j$ must not be in the same subset, when $c_{ij} = 0$ there simply is no indication of the repellent type. It was demonstrated in [6] that minimizing a sum of logarithmized pairwise conflicts,

$$\sum_i \sum_{\substack{k,l \\ e_k, e_l \in \chi_i}} -\log(1 - c_{kl}),$$

is with a small approximation identical to minimizing the metaconflict function, making it possible the map the optimization problem onto a neural network for neural optimization [2, 7].

In section 3 we will refine the frame of discernment and the proposition of $m_{ij}^-(\cdot)$ in order to make such a refined analysis possible.

In addition to this conflicting metalevel evidence from internal conflicts between belief functions, it is in many applications important to be able to handle attracting metalevel evidence from some external source stating that things do belong together, Figure 3. The analysis of this case is the contribution of the current paper.

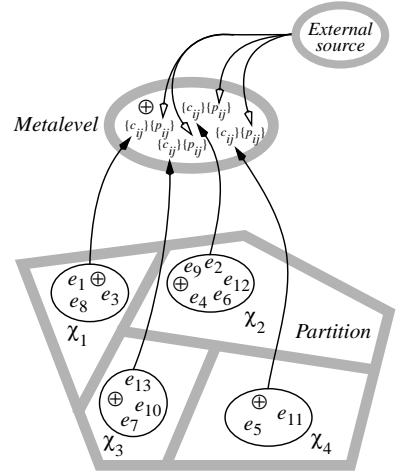

Figure 3: Conflict evidence from subsets and attracting evidence from an external source.

Let $m_{ij}^+(\cdot) = p_{ij}$, where $p_{ij}$ is a degree of attraction, be such an external metalevel evidence. When $p_{ij} = 1$, $e_i$ and $e_j$ must be in the same subset, when $p_{ij} = 0$ we have no indication of the attracting type.

Of course, we can also have external conflicting metalevel evidence. It is then combined with $m_{ij}^-(\cdot)$, and henceforth we will use $m_{ij}^-(\cdot)$ as the combined result if such external evidence is present.

## 3 Evidence level

Looking at a series of single subset problems, our frame of discernment for the metalevel of each subset $\chi_a$ was initially represented as $\Theta_a = \{AdP, \neg AdP\}$, [4]. It is here refined to

$$\Theta_a = \{\forall j. e_j \in \chi_a\} \cup \{e_j \notin \chi_a\}_{j=1}^{|\chi_a|},$$

where "adequate partition" AdP is refined to the proposition $\forall j. e_j \in \chi_a$, that each belief function $e_j$ placed in subset $\chi_a$ actually belongs to $\chi_a$. On the other hand, "not adequate partition" $\neg AdP$ is refined to a set of $|\chi_a|$ propositions $e_j \notin \chi_a$, each stating that a particular belief function is misplaced.

Thus, $|\Theta_a| = 1 + |\chi_a|$, where $|\chi_a|$ is the number of pieces of evidence in $\chi_a$.

Let us assign values to all conflicting and attracting pieces of metalevel evidence. However, we will not combine the attracting and conflicting evidence regarding each pair here on the evidence level as this result is currently not our concern.

## 3.1 Conflicting evidence: $m_{ij}^-(\cdot)$.

For each pair of belief functions we may receive a conflict. We interpret this as a piece of metalevel evidence indicating that the two belief functions do not belong to the same subset,

$$m_{ij}^-(\forall a. e_i \in \chi_a \Rightarrow e_j \notin \chi_a) = c_{ij},$$
$$m_{ij}^-(\Theta) = 1 - c_{ij}.$$

Here, we simply state that if $e_i$ belongs to a subset $\chi_a$ then $e_j$ must not belong to the same subset. Instead, we could have made a disjunction of two different propositions $[(e_i \in \chi_a \Rightarrow e_j \notin \chi_a) \vee (e_j \in \chi_a \Rightarrow e_i \notin \chi_a)]$ where $i \leftrightarrow j$ is permuted in the second term, but this is unnecessary and redundant information because of symmetry.

The metalevel evidence may be simplified to

$$m_{ij}^-(e_i \vee e_j \notin \chi_a) = c_{ij},$$
$$m_{ij}^-(\Theta) = 1 - c_{ij},$$

since

$$\forall a. e_i \in \chi_a \Rightarrow e_j \notin \chi_a = \forall a. \neg e_i \in \chi_a \vee e_j \notin \chi_a$$
$$= \forall a. e_i \notin \chi_a \vee e_j \notin \chi_a = \forall a. e_i \vee e_j \notin \chi_a$$
$$= e_i \vee e_j \notin \chi_a$$

by implication replacement and dropping universal quantifiers.

We calculate $m_{ij}^-$ for all pairs (*ij*).

## 3.2 Attracting evidence: $m_{ij}^+(\cdot)$.

In addition we may also have attracting evidence brought in externally. Such a piece of metalevel evidence is interpreted as the negation of the previous proposition, i.e., that the two pieces of evidence belong to the same cluster,

$$m_{ij}^+(\neg \forall a. e_i \in \chi_a \Rightarrow e_j \notin \chi_a) = p_{ij},$$
$$m_{ij}^+(\Theta) = 1 - p_{ij}.$$

Simplified to

$$m_{ij}^+(e_i \wedge e_j \in \chi_a) = p_{ij},$$
$$m_{ij}^+(\Theta) = 1 - p_{ij},$$

since

$$\neg \forall a. e_i \in \chi_a \Rightarrow e_j \notin \chi_a = \exists a. \neg (e_i \in \chi_a \Rightarrow e_j \notin \chi_a)$$
$$= \exists a. \neg (\neg e_i \in \chi_a \vee e_j \notin \chi_a)$$
$$= \exists a. \neg (\neg e_i \in \chi_a \vee \neg e_j \in \chi_a)$$
$$= \exists a. \neg \neg (e_i \in \chi_a \wedge e_j \in \chi_a) = \exists a. e_i \in \chi_a \wedge e_j \in \chi_a$$
$$= \exists a. e_i \wedge e_j \in \chi_a = e_i \wedge e_j \in \chi_a$$

by bringing in negation, implication replacement and dropping of universal quantifiers.

We calculate $m_{ij}^+$ for all pairs (*ij*).

Having assigned values to all conflicting and attracting metalevel evidence regarding every pair of belief functions we take the analysis to the cluster level.

## 4 Cluster level

At the cluster level we use the evidence derived in the previous level. We also use the same frame of discernment. Let us separately combine all conflicting $\{m_{ij}^-\}$ and all attracting evidence $\{m_{ij}^+\}$ for each cluster.

## 4.1 Combine all conflicting evidence within each cluster

Let us combine $\forall i, j, a. \oplus \{m_{ij}^- \mid m_{ij}^- \in \chi_a\}$, i.e., all conflicting metalevel evidence within each subset where $m_{ij}^-(e_i \vee e_j \notin \chi_a) = c_{ij}$, Section 3.1.

In [5] we refined the proposition $\neg$AdP separately for each cluster $\chi_a$ to $\exists j. e_j \notin \chi_a$, i.e., that there is at least one belief function misplaced in the subset.

Consequently, from the result of the above combination we have,

$$m_{\chi_a}^-(\neg\text{AdP}) = m_{\chi_a}^-(\exists j.e_j \notin \chi_a)$$
$$= 1 - \prod_{(ij) \in \chi_a}(1 - c_{ij}),$$
$$m_{\chi_a}^-(\Theta) = 1 - m_{\chi_a}^-(\neg\text{AdP}).$$

We calculate $m_{\chi_a}^-$ for all subsets $\chi_a$. This is the conflicting metalevel evidence derived at the cluster level.

In addition, this piece of evidence $m_{\chi_a}^-$ with proposition $\neg\text{AdP}$ may at the next level be refined as $\chi_a \notin \chi$, where $\chi$ is the set of all subsets. That is, the same conflict that on the cluster level is interpreted as if there is at least one belief function that does not belong to $\chi_a$, will on the partition level be interpreted as if $\chi_a$ (i.e., with all its content) does not belong to $\chi$. This will be useful at the partition level when combining all $m_{\chi_a}^-$ for different subsets $\chi_a$.

## 4.2 Combine all attracting evidence within each cluster

Similarly to the previous section we begin by combining all attracting metalevel evidence within each individual subset, $\forall i, j, a. \oplus \{m_{ij}^+ | m_{ij}^+ \in \chi_a\}$, where $m_{ij}^+$ was derived as $m_{ij}^+(e_i \wedge e_j \in \chi_a) = p_{ij}$ in Section 3.2.

For attracting metalevel evidence we refine AdP as the negation of the refinement of $\neg\text{AdP}$. We have,

$$\text{AdP} = \neg\neg\text{AdP} = \neg\exists j.e_j \notin \chi_a$$
$$= \forall j.\neg e_j \notin \chi_a = \forall j.e_j \in \chi_a.$$

We need to calculate the support for an adequate partition from all attracting evidence $m_{\chi_a}^+(\text{AdP})$ in each subset $\chi_a$. Thus, we will sum up the contribution from all intersections corresponding to a proposition that a conjunction of all pieces of evidence placed in the cluster actually belongs to the subset in question, i.e., $\wedge \{e_j \in \chi_a\}_{j=1}^{|\chi_a|}$.

From the combination of all $\{m_{ij}^+\}$ we have,

$$m_{\chi_a}^+(\text{AdP}) = m_{\chi_a}^+(\forall j.e_j \in \chi_a)$$
$$= \sum_{I \subseteq P_{|\chi_a|} | M_I \equiv N_{|\chi_a|}} \prod_I p_{ij} \prod_{P_{|\chi_a|} - I} (1 - p_{ij}),$$
$$m_{\chi_a}^+(\Theta) = 1 - m_{\chi_a}^+(\text{AdP}),$$

where $P_{|\chi_a|} = \{(ij) | 1 \leq i < j \leq |\chi_a|\}$ is a set of all pairs of ordered numbers $\leq |\chi_a|$, $M_I = \{i | \exists p.(ip) \vee (pi) \in I\}$ is the set of all numbers in the pairs of $I$, and $N_{|\chi_a|} = \{1, ..., |\chi_a|\}$ is the set of all numbers $\leq |\chi_a|$.

We calculate $m_{\chi_a}^+$ for all subsets $\chi_a$. This is the attracting metalevel evidence on the cluster level.

In addition AdP may on the next level be refined as $\chi_a \in \chi$. This will be useful at the partition level when combining all $m_{\chi_a}^+$ from the different subsets.

## 5 Partition level

The partition level is where all things come together, Figure 1. First, we combine all conflicting metalevel evidence from the subsets, $m_{\chi_a}^-$, Section 5.1. Secondly, we combine all attracting metalevel evidence from the same subsets, $m_{\chi_a}^+$, Section 5.2. Finally, we combine the conflicting and attracting metalevel evidence (in Section 5.3).

However, before we start, let us notice that on the partition level we do not reason about misplaced belief functions. Instead, we reason about the different parts of the partition (i.e., the subsets), and whether each of the subsets can make up part of an adequate partition. For this reason we should represent the frame of discernment differently than on previous levels.

The frame of discernment on the partition level $\Theta = \{\text{AdP}, \neg\text{AdP}\}$ is refined as

$$\Theta = \{\forall a.\chi_a \in \chi\} \cup \{\chi_a \notin \chi\}_{a=1}^{|\chi|},$$

where "adequate partition" AdP is refined to a the proposition $\forall a.\chi_a \in \chi$, stating that every subset $\chi_a$ does make up part of an adequate partition. On the other hand, "not adequate partition" $\neg\text{AdP}$ is refined to a set of $|\chi|$ propositions $\chi_a \notin \chi$, each stating that a particular subset does not make up part of an adequate partition.

Thus, the size of the frame is $|\Theta| = 1 + |\chi|$, where $|\chi|$ is the number of subsets in $\chi$.

## 5.1 Combine all conflicting evidence at the partition level

We begin by combining $\forall a. \oplus \{m_{\chi_a}^-\}$, i.e., all conflicting metalevel evidence from the subsets $\chi_a$ that we derived in Section 4.1.

Let us then refine the proposition $\neg\text{AdP}$ of $m_{\chi_a}^-(\neg\text{AdP})$ such that $\neg\text{AdP} = \exists a.\chi_a \notin \chi$, i.e., that there is at least one subset that does not make up part of an adequate partition.

From the combination of all $\{m_{\chi_a}^-\}$ we have,

$$m_\chi^-(\neg\text{AdP}) = m_\chi^-(\exists a.\chi_a \notin \chi)$$
$$= 1 - \prod_a [1 - m_{\chi_a}^-(\chi_a \notin \chi)]$$
$$= 1 - \prod_a [1 - m_{\chi_a}^-(\neg\text{AdP})]$$
$$= 1 - \prod_a \prod_{(ij) \in \chi_a} (1 - c_{ij}),$$
$$m_\chi^-(\Theta) = 1 - m_\chi^-(\neg\text{AdP}).$$

This is the conflicting metalevel evidence at the partition level.

## 5.2 Combine all attracting evidence at the partition level

Let us combine all attracting metalevel evidence $\forall a. \oplus \{m_{\chi_a}^+\}$, derived in Section 4.2.

For attracting metalevel evidence at the partition level we refine the proposition AdP of $m_\chi^+(\text{AdP})$ as the negation of the refinement for $\neg\text{AdP}$ at this level,

$$\text{AdP} = \neg\neg\text{AdP} = \neg\exists a.\chi_a \notin \chi$$
$$= \forall a.\neg\chi_a \notin \chi = \forall a.\chi_a \in \chi.$$

From the combination of all $\{m_{\chi_a}^+\}$ we find,

$$m_\chi^+(\text{AdP}) = m_\chi^+(\forall a.\chi_a \in \chi)$$
$$= \prod_a m_{\chi_a}^+(\chi_a \in \chi)$$
$$= \prod_a m_{\chi_a}^+(\text{AdP})$$
$$= \prod_a \sum_{I \subseteq P_{|\chi_a|} | M_I \equiv N_{|\chi_a|}} \prod_I p_{ij} \prod_{P_{|\chi_a|} - I} (1 - p_{ij}),$$
$$m_\chi^+(\Theta) = 1 - m_\chi^+(\text{AdP}).$$

This is the attracting metalevel evidence at the partition level.

## 5.3 Combine conflicting and attracting evidence

As the final step on the partition level (Figure 1) we combine all already combined conflicting evidence (Section 5.1) with all already combined attracting evidence (Section 5.2), $m_\chi = m_\chi^+ \oplus m_\chi^-$. We receive,

$$m_\chi(\text{AdP}) = m_\chi(\forall a.\chi_a \in \chi)$$
$$= m_\chi^+(\forall a.\chi_a \in \chi)m_\chi^-(\Theta)$$
$$= m_\chi^+(\text{AdP})m_\chi^-(\Theta),$$
$$m_\chi(\neg\text{AdP}) = m_\chi(\exists a.\chi_a \notin \chi)$$
$$= m_\chi^+(\Theta)m_\chi^-(\exists a.\chi_a \notin \chi)$$
$$= m_\chi^+(\Theta)m_\chi^-(\neg\text{AdP}),$$
$$m_\chi(\Theta) = m_\chi^+(\Theta)m_\chi^-(\Theta),$$
$$m_\chi(\varnothing) = m_\chi^+(\forall a.\chi_a \in \chi)m_\chi^-(\exists a.\chi_a \notin \chi)$$
$$= m_\chi^+(\text{AdP})m_\chi^-(\neg\text{AdP}).$$

With a conflict $m_\chi(\varnothing)$, since

$$(\forall a.\chi_a \in \chi) \wedge (\exists a.\chi_a \notin \chi) = \texttt{False}.$$

This is the amount of support awarded to the proposition that we have an "adequate partition" $m_\chi(\text{AdP})$, and awarded to the proposition that we do not have an "adequate partition" $m_\chi(\neg\text{AdP})$, respectively, when taking everything into account.

# 6 Weighting by information content

In order to find the best partition we might want to maximize $m_\chi(\text{AdP})$. However, in the special case when there is no positive metalevel evidence then $m_\chi(\text{AdP}) = m_\chi^+(\text{AdP}) = 0$. Alternative, we might like to minimize $m_\chi(\neg\text{AdP})$. This is what was done in [4] when only negative metalevel evidence was available. However, here we also have a special case when there is no negative metalevel evidence. Then, $m_\chi(\neg\text{AdP}) = m_\chi^-(\neg\text{AdP}) = 0$. The obvious solution is to minimize a function of $m_\chi(\neg\text{AdP})$ and $-m_\chi(\text{AdP})$. In doing this, we want to give each term a weighting corresponding to the information content of all conflicting and all attracting metalevel evidence, respectively. This is done in order to let each part have an influence corresponding to its information content.

Thus, let us minimize a metaconflict function

$$Mcf = \alpha[1 - m_\chi(\text{AdP})] + (1 - \alpha)m_\chi(\neg\text{AdP}),$$

$0 \leq \alpha \leq 1$, where $\alpha = 0$ when all $p_{ij} = 0$, and $\alpha = 1$ when all $c_{ij} = 0$.

Let

$$\alpha = \frac{H(m_{\chi_0}^+)}{H(m_{\chi_0}^+) - H(m_{\chi_0}^-)},$$

where $H(m)$ is the expected value of the entropy $-\log_2[m(A)/|A|]$. $H(m)$ is called the average total uncertainty [3], measuring both scattering and nonspecificity, and may be written as the sum of Shannon entropy, $G(\cdot)$, and Hartley information, $I(\cdot)$,

$$\begin{aligned} H(m) &= G(m) + I(m) \\ &= -\sum_{A \in \Theta} m(A)\log_2[m(A)] \\ &\quad + \sum_{A \in \Theta} m(A)\log_2(|A|). \end{aligned}$$

Here, $m_{\chi_0}^+$ and $m_{\chi_0}^-$ are calculated on the cluster level, as if all evidence is put into one large imaginary cluster $\chi_0$.

## 6.1 Entropy of conflicting metalevel evidence $H(m_{\chi_0}^-)$

First, we combine $\forall i, j. \oplus \{m_{ij}^-\}$, i.e., all conflicting metalevel evidence, taking no account of which subset the different $m_{ij}^-$ actually belongs to.

In this combination all intersections in the combined result are unique. Thus, the number of focal elements are equal to the number of intersections as no two intersections add up. Calculating the average total uncertainty of all conflicting metalevel evidence $H(m_{\chi_0}^-) = G(m_{\chi_0}^-) + I(m_{\chi_0}^-)$ is then rather simple,

$$G(m_{\chi_0}^-) = -\sum_{J \subseteq P_n} m_{\chi_0}^-(\wedge\{e_i \vee e_j | (ij) \in J\} \notin \chi_0)$$
$$\cdot \log_2[m_{\chi_0}^-(\wedge\{e_i \vee e_j | (ij) \in J\} \notin \chi_0)],$$

where $P_n = \{(ij) | 1 \leq i < j \leq n\}$ is a set of pairs $(ij)$, and $n$ is the number of belief functions, with

$$\forall J \subseteq N | |J| > 1. m_{\chi_0}^-(\wedge\{e_i \vee e_j | (ij) \in J\} \notin \chi_0)$$
$$= \prod_{I \subseteq P} c_{ij} \prod_{P-I} (1 - c_{ij}).$$

The Hartley information is calculated as

$$I(m_{\chi_0}^-) = -\sum_{J \subseteq P_n} m_{\chi_0}^-(\wedge\{e_i \vee e_j | (ij) \in J\} \notin \chi_0)$$
$$\cdot \log_2(|J|).$$

## 6.2 Entropy of attracting metalevel evidence $H(m_{\chi_0}^+)$

Similarly, we combine $\forall i, j. \oplus \{m_{ij}^+\}$, i.e., regardless of which subset the $m_{ij}^+$'s actually belongs to.

When calculating $H(m_{\chi_0}^+) = G(m_{\chi_0}^+) + I(m_{\chi_0}^+)$ the Shannon entropy may be calculated as

$$G(m_{\chi_0}^+) = -\sum_{J \subseteq N | |J| > 1} m_{\chi_0}^+(\wedge\{e_j | j \in J\} \in \chi_0)$$
$$\cdot \log_2[m_{\chi_0}^+(\wedge\{e_j | j \in J\} \in \chi_0)],$$

where $N = \{1, ..., n\}$, and $n$ is the number of belief functions, and

$$\forall J \subseteq N | |J| > 1 . m_{\chi_0}^+ ( \wedge \{e_j | j \in J\} \in \chi_0)$$
$$= \sum_{I \subseteq P | M_I \equiv J} \prod_I p_{ij} \prod_{P-I} (1 - p_{ij}),$$
$$m_{\chi_0}^+(\Theta) = \prod_{(ij)} (1 - p_{ij}),$$

where $P = \{(ij) | 1 \leq i < j \leq n\}$.

With Hartley information calculated as

$$I(m_{\chi_0}^+) = - \sum_{J \subseteq N | |J| > 1} m_{\chi_0}^+ ( \wedge \{e_j | j \in J\} \in \chi_0)$$
$$\cdot \log_2(|N - J| + 1).$$

## 7 Clustering belief functions

The best partition of all belief functions is found by minimizing

$$Mcf = \alpha[1 - m_\chi(\text{AdP})] + (1 - \alpha)m_\chi(\neg\text{AdP})$$

over all possible partitions. For a small number of belief functions this may be achieved through iterative optimization, but for a larger number of belief functions we need a method with a lower computational complexity, e.g., some neural clustering method similar to what was done in the case with only conflicting metalevel evidence [2].

## 8 Conclusions

We have extended the methodology for clustering belief function from only being able to manage conflicting information [1–2, 7] to also being able to handle attracting information. This is important in many practical applications within information fusion.